\documentclass[letterpaper, 10 pt, conference]{ieeeconf}  

\IEEEoverridecommandlockouts                              

\usepackage{graphicx} 
\usepackage{balance}
\usepackage{booktabs}
\usepackage{graphics}
\usepackage{textcomp}
\usepackage{hyperref}
\let\labelindent\relax \usepackage{enumitem}
\usepackage{multirow}

\title{\LARGE \bf
\textit{robo-gym} -- An Open Source Toolkit for Distributed Deep Reinforcement Learning on Real and Simulated Robots
}

\author{Matteo Lucchi$^{*}$, Friedemann Zindler$^{*}$, Stephan M\"{u}hlbacher-Karrer, Horst Pichler 
\thanks{* These authors contributed equally.}
\thanks{All authors are with Joanneum Research -- Robotics, Klagenfurt am W\"orthersee, Austria, \tt{first.lastname@joanneum.at}}
}

\begin{document}

\maketitle
\thispagestyle{empty}
\pagestyle{empty}

\begin{abstract}



Applying Deep Reinforcement Learning (DRL) to complex tasks in the field of robotics has proven to be very successful in the recent years. 
However, most of the publications focus either on applying it to a task in simulation or to a task in a real world setup. 
Although there are great examples of combining the two worlds with the help of transfer learning, it often requires a lot of additional work and fine-tuning to make the setup work effectively.
In order to increase the use of DRL with real robots and reduce the gap between simulation and real world robotics, we propose an open source toolkit: \textit{robo-gym}\footnote{Source code and application videos are available at: \newline \url{https://sites.google.com/view/robo-gym}}. We demonstrate a unified setup for simulation and real environments which enables a seamless transfer from training in simulation to application on the robot. 
We showcase the capabilities and the effectiveness of the framework with two real world applications featuring industrial robots: a mobile robot and a robot arm. 
The distributed capabilities of the framework enable several advantages like using distributed algorithms, separating the workload of simulation and training on different physical machines as well as enabling the future opportunity to train in simulation and real world at the same time.
Finally, we offer an overview and comparison of \textit{robo-gym} with other frequently used state-of-the-art DRL frameworks.
\end{abstract}

\section{INTRODUCTION}

Traditionally, industrial robots have been operating in closed cells or warehouse areas with limited access. 
In most cases, these robots perform well-defined, repeated operations on standard objects without interacting with human operators.
Programming a robot is often a lengthy task that requires specialized knowledge of the machine's software. 
Recent trends in robotics aim to enable robots to work in dynamic, open environments co-occupied by humans, which present several new challenges. 
When working in these complex scenarios, a robot must be equipped with certain sensors that allow it to perceive its surroundings and the objects it has to interact with.
Integrating and exploiting sensor data for planning the robot's actions is not a trivial task.

Research has shown that applying DRL to solve complex robotics tasks is a promising solution to the shortcomings of traditional methods. 
%
Many existing frameworks and toolkits have been developed by researchers in the AI community to test and compare their algorithms on a set of very complex problems. 
The results obtained are very impressive, but the applications are mostly confined to the simulation world and are rarely transferred to the real world. 
Closing the gap between simulation and real world is an incredibly promising mission on which many researchers are currently working. 
However, DRL is a complex field of research that requires in-depth knowledge in several areas, which in our experience represents a barrier to entry for roboticists.


 \begin{figure}[!tb]
      \centering
      \includegraphics[width=0.99\columnwidth]{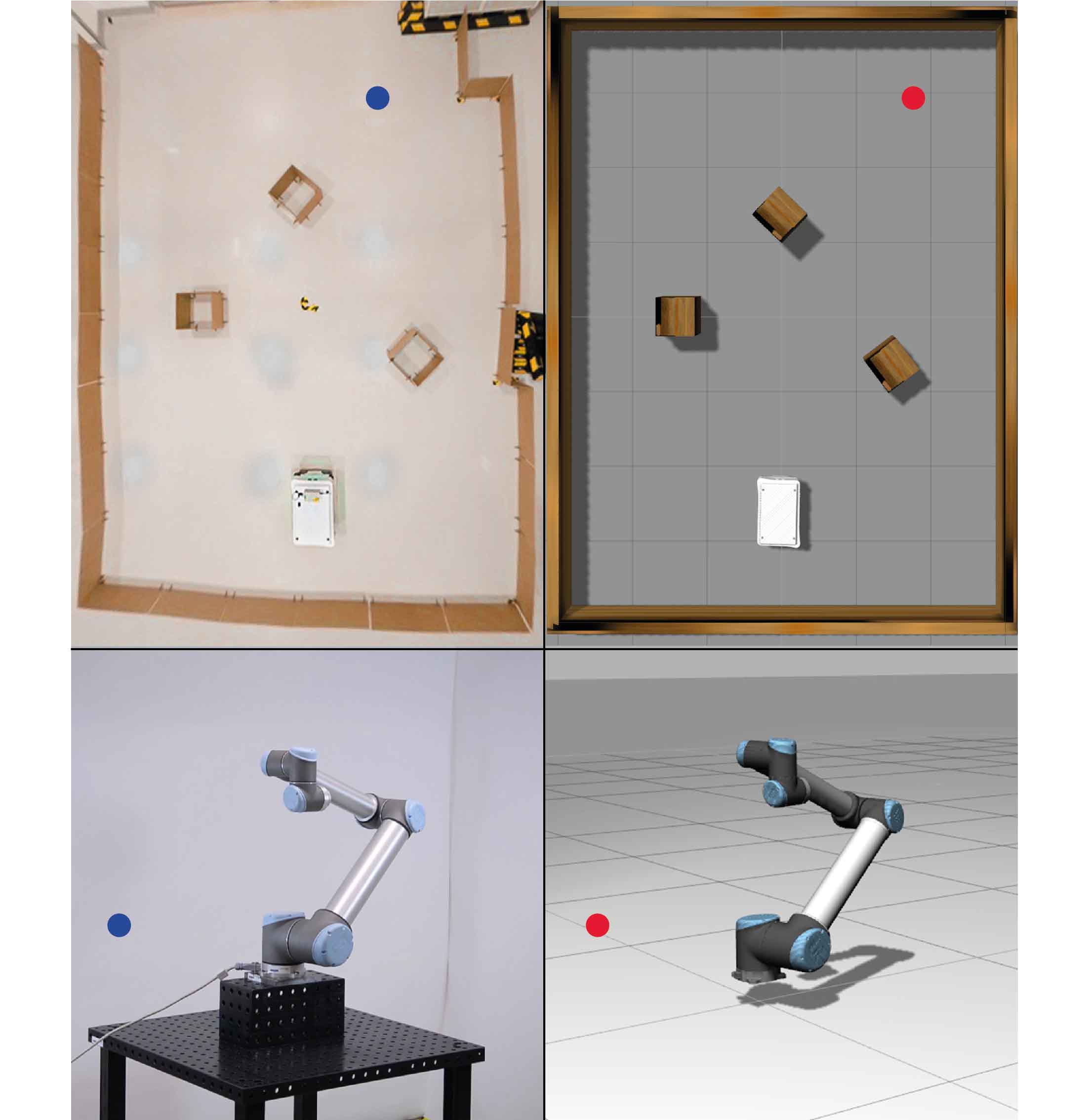}
      
      \caption{Industrial robot use case scenarios (left: real environment, right: simulation environment). Mobile navigation with obstacle avoidance of MiR100 on the top and end effector positioning of UR10 on the bottom.}
      \label{fig:use_case}
 \end{figure}

%
%
%
%
%

% ...

%
%
% ...




Our \textit{contribution}, the \textit{robo-gym} framework, is an open source toolkit for DRL on real and simulated robots and creates a bridge between communities.
By using a standardized interface based on OpenAI Gym,
we enable AI researchers to test their algorithms on simulated and real world problems involving industrial robots with little or no knowledge in the robotics field. 
On the other hand, robotic researchers are able to focus on the integration of new robots, sensors and tasks, while exploiting many of the open source implementations of DRL algorithms using the OpenAI Gym interface (e.g. Stable Baselines \cite{stable-baselines}). 

During the implementation of the proposed framework we encountered several issues when dealing with real world systems, and while there are examples of applications tested on real robots, only few works share details about the hardware setup and interfaces. 
To help the researchers set up similar tasks, we provide examples of two industrial use cases with a UR10 collaborative robot arm and MiR100 mobile robot (see Figure~\ref{fig:use_case}).

We built \textit{robo-gym} to be able to quickly develop and train new applications on our own hardware, without having to be tied to cloud services providers, and to deploy them in industrial use cases.  
We provide this tool to the community with the goal to accelerate research in this field; furthermore, we commit to actively maintain the framework and continuously extend it with new robot models, sensors and tasks.





The remainder of the paper is structured as follows: Section~\ref{sec:relatedwork} gives an overview of related work. Section~\ref{sec:framework} describes each component of the \textit{robo-gym} framework in detail, introduces how the elements operate together, and how to extend them. Section~\ref{sec:application} shows how we applied our framework on two proof-of-concept use cases. Section~\ref{sec:framework_evaluation} compares \textit{robo-gym} to other popular state-of-the-art frameworks and the conclusion is given in Section~\ref{sec:conclusions}. 

\section{Related Work}
\label{sec:relatedwork}

\subsection{Deep Reinforcement Learning in Robotics}




In robot arm manipulation, tasks are differentiated according to the given input. Some tasks use the robot's proprioceptive information, while others most often use visual data from RGB cameras or even a combination of the two.
Initial research solved diverse robot arm manipulation tasks mainly in simulation \cite{Gu2017}, \cite{Popov2017}, \cite{Huang}. 
Other approaches trained directly on the real robot, but this is difficult and requires to have a lot of constraints on the movements of the robots \cite{Gu2017}, \cite{Chebotar2017}, \cite{Mahmood2018a}, \cite{Levine2015}. 
A more recent work tried to combine the two domains by pre-training models in simulation and continuing learning in the real world \cite{Rusu2017}.
Subsequently, latest advancements use domain randomization to train models in simulation and deploy them on the real robots, with \cite{VanBaar2019a}, \cite{Tobin2017} or without \cite{Peng2018}, \cite{Matas2018}, \cite{Antonova2017}, \cite{Akkaya2019} some additional fine tuning involved.

In mobile robot navigation, DRL has also demonstrated its applicability, where it is very practical to map actions to large sensory data.
Renowned examples of problems solved both in simulation and in the real world are: mobile navigation with static \cite{Tai2017d} and dynamic \cite{Chiang2019} obstacle avoidance, decentralized multi robot collision avoidance \cite{Long2018d} and socially-compliant navigation in crowded spaces \cite{Chen2017}, \cite{Chen2019}. 





\smallskip

\subsection{Frameworks and Benchmarks} \label{exisiting_frameworks}

OpenAI Gym~\cite{openai2016} has become the de facto standard for benchmarking DRL algorithms; it includes a suite of robotics environments based on the MuJoCo simulation engine \cite{Todorov2012}, but it does not serve all the needs of the robotics community. 
As a consequence, several research groups and companies have tried to set a standard for developing and  benchmarking DRL applications in robotics.


The DeepMind Control Suite \cite{Tassa2018} aims at providing a benchmark for performance comparison of learning algorithms on exclusively simulated physics-control problems. 

The Surreal Robotics Suite \cite{Fan*2018} focuses on robotic manipulation.
It includes multiple simulated tasks with the Baxter robot. 

RLBench \cite{James2019} aims at providing a large-scale benchmark and learning environment specifically tailored for vision-guided manipulation research. 

The SenseAct framework \cite{Mahmood2018a} includes learning environments based on multiple real robots.  
The industrial robot environments are developed only for the real hardware and not in simulation.

The toolkit gym-gazebo2 \cite{Lopez2019}, which is based on ROS2 and Gazebo, comes with environments using both the real and the simulated MARA Robot formerly developed by Acutronic Robotics; it is the most closely related to our work. 


However, \textit{robo-gym} is the only framework that allows to train control policies on distributed simulations and to exploit them directly with commercially available robots. A more extensive comparison to  related frameworks is given in Section~\ref{sec:framework_evaluation}.


\smallskip

\subsection{Physics Engines}\label{physics_engines}

There is no clear preference when it comes to physics engines for robotics simulation, mostly due to the fact that each of the popular simulation platforms have strengths and weaknesses on different kind of problems.

MuJoCo \cite{Todorov2012} is well known for the accuracy of its contact and friction simulations and it has become very popular among the AI community. 
It is used by several frameworks and it is well suited for research on low level control of complex physical systems with a high number of degrees of freedom. 
On the negative side, it lacks integration with other tools commonly used by roboticists.
Furthermore, the software is proprietary and it has prohibitive licence costs. 

Coppelia Sim, formerly known as V-rep \cite{Rohmer2013}, and Gazebo \cite{Koenig2004} are both popular simulation platforms within the robotics community. 
Both of them can exploit different physics engines like Open Dynamics Engine (ODE) or Bullet. 
Coppelia Sim is proprietary although parts of the software are open source whereas Gazebo is completely open source.

The latter is a very popular choice for roboticists.
Some of the main benefits of using Gazebo are: the community support, the vast library of robots and sensors as well as the integration with the Robot Operating System (ROS).
Furthermore, the availability of ROS controllers and sensors plugins allow to have similar interfaces for the simulated and the real robots. 

\section{THE FRAMEWORK}
\label{sec:framework}


%

\begin{figure}[thpb]
    \centering
    \includegraphics[width=1\columnwidth]{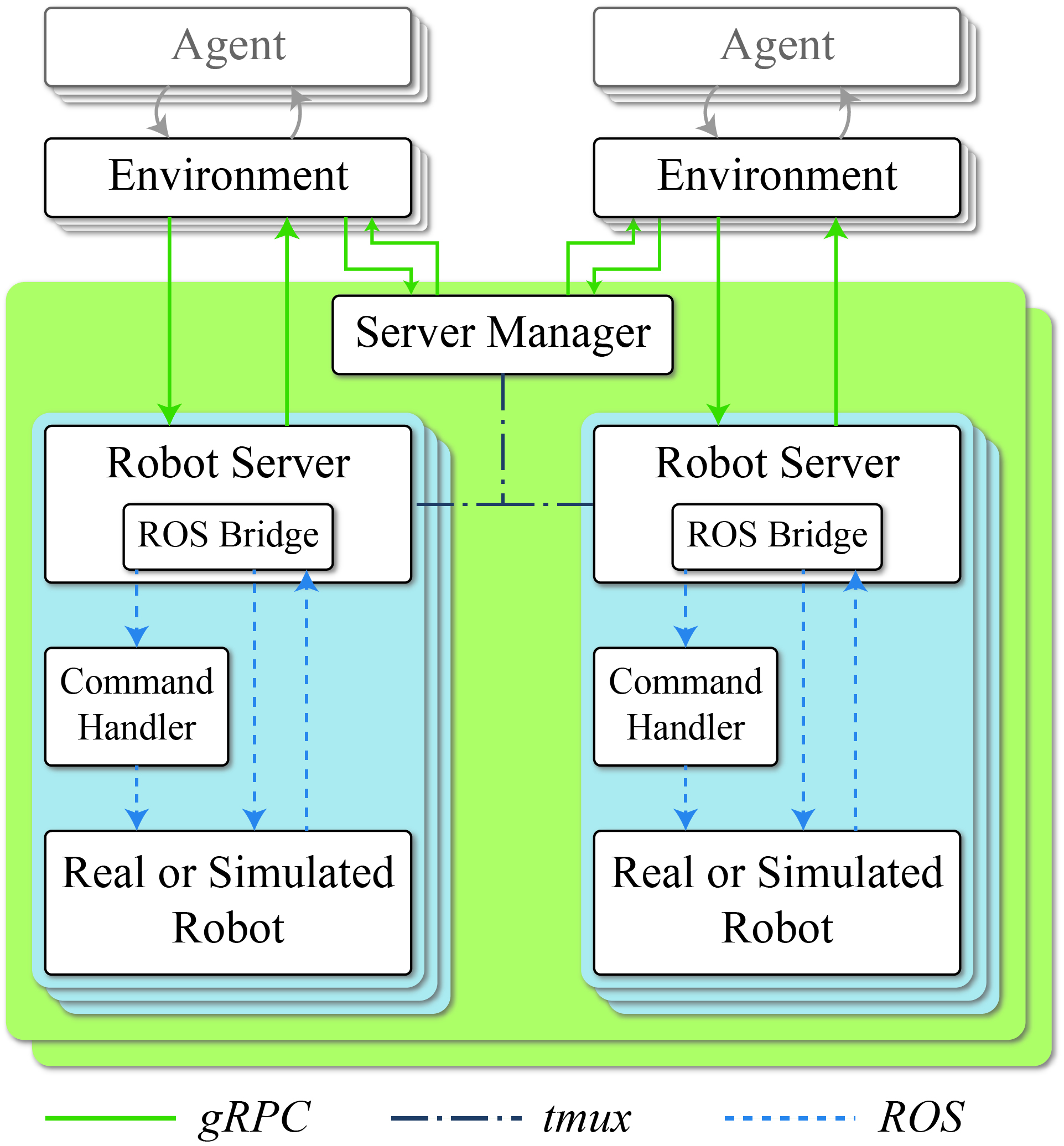}
    \caption{The \textit{robo-gym} framework.}
    \label{fig:framework}
 \end{figure}
 
 
\subsection{The Components}
The elements of the framework, depicted in Figure~\ref{fig:framework}, are introduced following a bottom-up approach, starting from the hardware layer building up to the interface to the Reinforcement Learning (RL) agent.  
\smallskip

\subsubsection{Real or Simulated Robot} 
This component includes the robot itself, the sensors, and the scene surrounding the robot. 

The interface to the robots and the sensors is implemented in ROS for both real and simulated hardware. The interface of the simulated robots is generally corresponding to the one of the real robots augmented with features that would be impractical or too expensive to match in the real world. An example is virtual collision sensors that detect any kind of collision of a robot link.
The simulated and the real robot must use the same controllers.

The simulated scenes are generated in Gazebo and are described using the SDFormat (SDF), an XML format. These can be created and manipulated in multiple ways: online, via API or GUI, and offline, by editing the SDF files.

\smallskip

\subsubsection{Command Handler}

Within the Markov Decision Process (MDP) framework, it is assumed that interactions between agent and environment take place at each of a series of discrete time steps. 
According to \cite{Sutton2018}, in such a system time does not advance between making an observation and triggering a subsequent action.
In a real-world system, however, time passes in a continuous manner.
It is therefore necessary to make some adjustments to the real world system so that its behavior gets closer to the one defined by the MDP framework.
The Command Handler (CH) implements these aspects.

As explained in \cite{Mahmood2018a}, the robot-actuation cycle time is the time between the individual commands sent to the robot controller and the action cycle time is the time between two subsequent actions generated from the agent. 
The action cycle time doesn't have to be the same as the robot-actuation cycle time, as the CH can repeatedly publish the same command for multiple robot-actuation cycles. 

The CH uses a queue with capacity for one command message. 
When the component receives a command message it tries to fill the queue with it. 
New elements get ignored until the element in the queue gets consumed.   
The CH continuously publishes command messages to the robot at the frequency required by its controller. 
If, at the moment of publishing, the queue is full, the CH retrieves the command, publishes it to the robot for the selected number of times and after that it empties the queue. 
In the opposite case, the handler publishes the command corresponding to an interruption of the movement execution. This corresponds to either zero velocities for mobile robots or preemption of the trajectory execution for robot arms. 

The framework's Command Handler supports the standard \textit{diff\_drive\_controller} and \textit{joint\_trajectory\_controller} from  ROS controllers \cite{ros_control}.
A wide range of robots can be controlled using these;
nevertheless, this component can be easily implemented for any other ROS controller.

\smallskip

\subsubsection{Robot Server} \label{robot_server}
It exposes a gRPC server that allows to interact with the robot through the integrated ROS bridge.

The first function of the server is to store updated information regarding the state of the robot, that can be queried at any time via a gRPC service call.
The robot's actuators and sensors constantly publish information via ROS.
The ROS Bridge collects the information from the different sources and stores it in a buffer as an array of values.
The actuators and the sensors update their information with different frequencies. 
The buffer is managed with a threading mechanism to ensure that the data delivered to the client is consistent and containing the latest information. 

The second function is to set the robot and the scene to a desired state. For instance, the user might want to set the joint positions of a robotic arm to a specific value when resetting the environment. 

Lastly, it provides a service to publish commands to the CH. 


\smallskip

\subsubsection{Environment}
This is the top element of the framework, which provides the standard OpenAI Gym interface to the agent. 
%
%
The main function of the Environment component is to define the composition of the state, the initial conditions of an episode and the reward function. 
In addition, it includes a gRPC stub which connects to the Robot Server to send actions, and to set or get the state of the robot and the scene. 

According to the framework provided by the Gym, environments are organized in classes, each constructed from a common base one. In addition, \textit{robo-gym} extends this setup with a different wrapper for either a real or a simulated robot. These wrappers differentiate regarding the constructor that is being called. 
In the case of the simulated robot environment, the argument for the IP address refers to the Server Manager, whereas in the case of the real robot environment it refers to the IP address of the Robot Server.
The Server Manager for simulated robots provides the address of the  Robot Server to which the Environment gRPC stub is then connected. 
On the other hand, in the case of the real robot environment, extra attention for the network configuration is needed to guarantee communication with the hardware. Furthermore, environment failures and eventual emergency stops must be inspected by a human operator. 
As a consequence, the Server Manager is currently not employed when using real robots and the Environment gRPC stub is connected directly to the Robot Server, which is started manually.


\smallskip
\subsubsection{Server Manager}
It is the orchestrator of the Robot Servers, it exposes gRPC services to spawn, kill, and check Robot Servers. 
When used with simulated robots it handles the robot simulations as well. 

Each cluster of Robot Server, CH and real or simulated robot runs on an isolated ROS network. 
To achieve this, the Server Manager launches each cluster in an isolated shell environment handled with the help of tmux\footnote{\url{https://github.com/tmux/tmux}}. 

This component implements error handling features to automatically restart the Robot Server and the robot simulation in case of:
\begin{itemize}
    \item an error in the connection to the Robot Server
    \item an exceeded deadline when calling a Robot Server service
    \item a non responding simulation 
    \item data received from simulation out of defined bounds
    \item a manual request of simulation restart
\end{itemize}


\smallskip

\subsection{The Process}

This subsection focuses on the run time behavior of the framework.
The most critical process that needs to be established is the one behind the call of a step in the environment. 
A RL agent uses $S_i,A_i,R_i,S_{i+1}$ tuples to train on the given environment, where $S$ is the state of the environment at different time steps, $A$ is the action taken in the environment and $R$ is the reward received. 
We will refer to the time necessary for the learning algorithm to generate the action as \textit{action generation time}.
Furthermore, we define the \textit{sleep time} as the difference between the action cycle time and the action generation time. 
\begin{figure}[thpb]
    \centering
    \includegraphics[width=1\columnwidth]{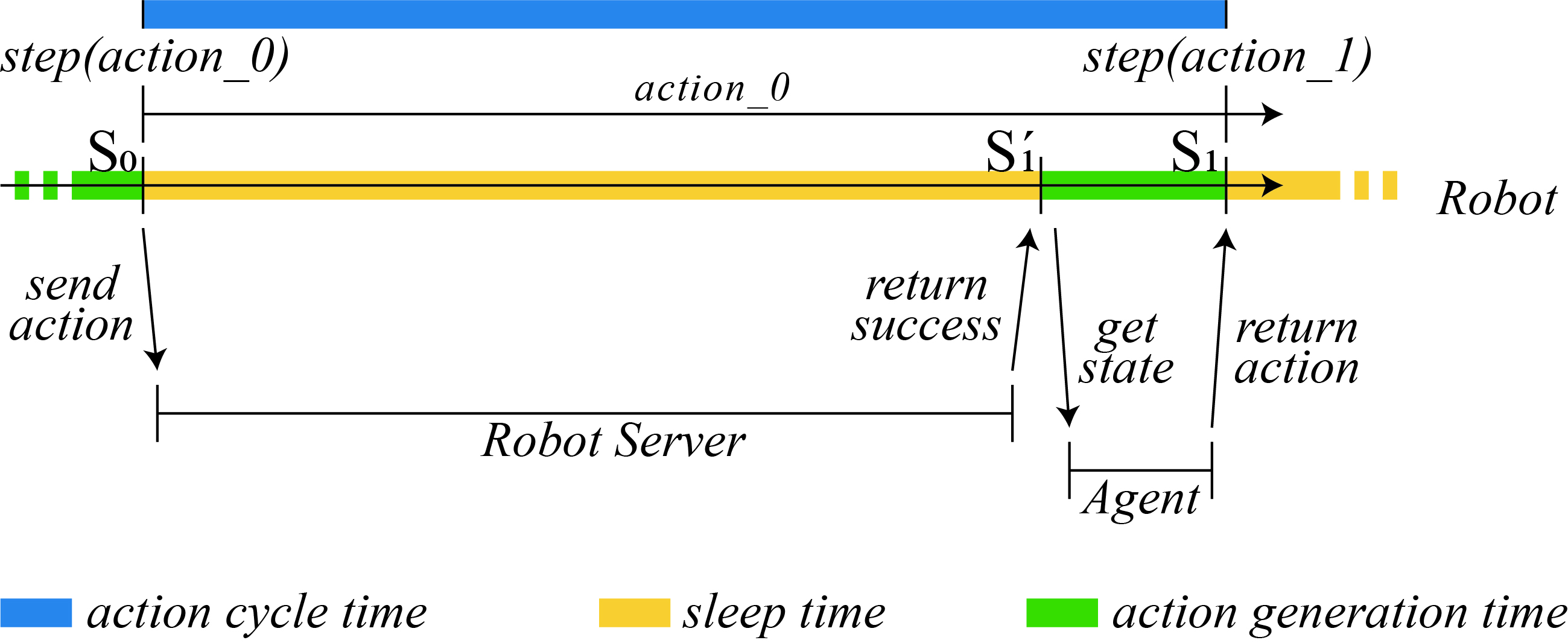}
    \caption{Process timeline of a step taken in the environment. }
    \label{fig:timeline_reg}
 \end{figure}
 
As shown in Figure \ref{fig:timeline_reg}, when calling a step in the environment the gRPC stub of the Environment component calls the Robot Server's service to send an action to the robot.
The Robot Server then publishes the desired command to the robot through the ROS bridge. Afterwards, it waits for the action execution time before returning the result of the service call. 
If no exceptions were raised during the execution, the gRPC stub of the Environment receives a positive feedback. 
Only after receiving the feedback, the Environment's gPRC stub queries the Robot Server's service to get the latest state of the robot. 
The action is generated based on the state  $S_1'$ different from the state $S_1$ at which the action is actually executed. 
This is unavoidable for real world systems and it highlights the importance of minimizing delays throughout the framework \cite{Mahmood2018a}.
Shorter action generation times allow to have finer and smoother control of the robot. 
The exact reference times for the two applications are further discussed in Section~\ref{real_world_setup}.

To distribute the computational efforts of the training process it is possible to run the framework across different PCs. 
The Robot Server, CH, and real or simulated robot clusters can be distributed across any PC connected to the same network. 
To start the framework it is sufficient to start a Server Manager on every PC and register its IP address. 
Although this has not been tested yet, it is also possible to train on real and simulated robots at the same time, due to the modular architecture based on gRPC.

\subsection{Extending the Framework}
\subsubsection{Extending Robotic Hardware}
New robot models and sensors can be easily integrated. In general for each different robot model a specific Robot Server, CH and Real or Simulated Robot are required. However, these can be implemented with a minor effort by adapting the components provided with the framework's code. New sensors must be integrated in the Robot Server in order to have the additional data forwarded to the Environment.
\smallskip
\subsubsection{Creating new Tasks}
When creating a new task, the only restrictions are those imposed by the simulator used. Thus, starting with a widely adopted simulator as Gazebo facilitates the process; since a large library of scenes and models has been already developed by the community. 
\smallskip
\subsubsection{Using Other Real World Systems or Simulators}

The two main reasons behind the use of gRPC as a communication layer are that it is open source and that it comes with libraries for multiple programming languages: C/C++, C\#, Dart, Go, Java, Node.js, Objective-C, PHP, Python and Ruby.
Thanks to the latter, it is possible to implement Robot Servers for any real world robot controller that provides an API in one of the supported languages. 
This is valid for robot simulators as well, since any simulator that provides an API in one of the gRPC supported languages could be integrated in the framework.
Nevertheless, we encourage the users to use the ROS framework and Gazebo when available. 

It is possible to use the existing framework with simulated robots in Gazebo using a different physics engine.
In the provided environments, the default physics engine, ODE, was used.
Nevertheless, it is up to the user to select the physics engine best suited to the given requirements.

\section{APPLICATION} \label{sec:application}

To exhibit the flexibility of the framework and to prove its usefulness we implemented two applications based on two different types of industrial robots. 

The first application features the MiR100, a differential drive mobile robot with a maximum payload of 100 kg. 
This robot is widely adopted in industry and research and it can be employed for a number of different tasks, due to the multiple extensions built by third party companies. 
Furthermore, the backbone of the robot is based on ROS making it straight forward to interact with it using the ROS framework. 

The second application features a UR10, a collaborative industrial robot with a maximum payload of 10 kg and a 1300 mm reach.
The choice of the robot was motivated by its popularity in industry and research as well as the availability of a ROS driver.

The RL agent could successfully solve the given task for both of the proposed applications. 
After the training process was completed, the agent was then deployed on the real robots without any further training. At least in these two simple tasks it can be observed that the trained agent can be applied with a similar success rate in the real world.


\subsection{Problem Description}

The initial release of \textit{robo-gym} provides two environments showcasing a navigation task with the MiR100 and a positioning task with the UR10, two common industrial use cases shown in Figure~\ref{fig:use_case}.
\smallskip

\subsubsection{Mobile navigation with obstacle avoidance of MiR100} \label{navigation_task}

In this environment, the task of the mobile robot is to reach a target position without touching the obstacles on the way. 

In order to detect obstacles, the MiR100 is equipped with two laser scanners, which provide distance measurements in all directions on a 2D plane.
At the initialization of the environment the target is randomly placed on the opposite side of the map with respect to the robot's position. 
Furthermore, three cubes, which act as obstacles, are randomly placed in between the start and goal positions. The cubes have an edge length of 0.5 m, whereas the whole map measures 6x8 m.

The observations consist of 20 values. 
The first two are the polar coordinates of the target position in the robot's reference frame.
The third and the fourth value are the linear and angular velocity of the robot.
The remaining 16 are the distance measurements received from the laser scanner distributed evenly around the mobile robot. These values were downsampled from 2*501 laser scanner values to reduce the complexity of the learning task. %

The action is composed of two values: the target linear and angular velocity of the robot. 

The base reward that the agent receives at each step is proportional to the variation of the two-dimensional Euclidean distance to the goal position.
Thus, a positive reward is received for moving closer to the goal, whereas a negative reward is collected for moving away.
In addition, the agent receives a large positive reward for reaching the goal and a large negative reward in case of collision.

\smallskip

\subsubsection{End effector positioning of UR10} \label{positioning_task}

The goal in this environment is for the robotic arm to reach a target position with its end effector.

This task is similar to UR5 Reacher~\cite{Mahmood2018a}, but with less constraints on the initial and final conditions.
The target end effector positions are not confined inside a small boundary box, but are uniformly distributed across a semi-sphere of radius 1200 mm, which is close to the full working area of the UR10. 
Potential target points generated within the singularity areas of the working space are discarded. 
In addition, the starting position is not the middle of the boundary box, but a random robot configuration. 

The observations consist of 15 values: the spherical coordinates of the target with the origin in the robot's base link, the six joint positions and the six joint velocities.

The robot uses position control; therefore, an action in the environment consists of six normalized joint position values.

The reward function is similar to the one of \textit{Problem 1} with the difference that the Euclidean distance is calculated in the three-dimensional space. 
Both self collisions and collisions with the ground are taken into account and punished with a negative reward and termination of the episode.

\smallskip

%

\subsection{The Learning Algorithm}
To showcase a proof of concept regarding the learning as well as the distribution capabilities within the framework, an implementation of Distributed Distributional Deep Deterministic Policy Gradients (D4PG) \cite{Barth-Maron2018} was chosen. This includes the proposed extensions of n-step returns, prioritized experience replay \cite{Horgan2018}, \cite{Barth-Maron2018} and a critic value function modeled as a categorical distribution \cite{Bellemare2017}, \cite{Barth-Maron2018}. Furthermore, D4PG has shown state-of-the-art performance in continuous-control tasks \cite{Tassa2018a}.

Hyperparameters were chosen according to the proposed benchmarks for DDPG and D4PG in \cite{Tassa2018a} with only minor changes. 
The values can be found together with the application videos on the framework's web page.




\subsection{The Hardware Setup}

\subsubsection{Computer Setup for training}\label{computer_setup}

Both of the models have been trained using 21 instances of the environment, 20 for actual learning running on one PC (36 CPU cores) and one
for supervision of the learning process running on another PC (4 CPU cores).
The learning algorithm was running on a third computer (1 NVIDIA Tesla P100 GPU + 16 CPU cores). 

\smallskip
\subsubsection{Real World Setup}\label{real_world_setup}

For the real world experiments of \textit{Problem 1} an area, resembling the one used in simulation, was reproduced in our laboratory.
Standing barriers were utilized to delimit the area and to create obstacles.
During the tests, the obstacles' positions have been changed every 10 episodes. 
The RL agent, the Environment, the Robot Server and the CH were running on a PC connected via Wi-Fi to the MiR100's network.  
The robot-actuation cycle time and the action cycle time were both 100 ms.

For \textit{Problem 2} the UR10 has been installed on a welding table. 
The RL agent, the Environment, the Robot Server, the CH and the ROS Driver\footnote{\url{https://github.com/UniversalRobots/Universal_Robots_ROS_Driver}} were running on a PC connected via Ethernet to the UR10's controller. 
The robot-actuation cycle time was 8 ms whereas the action cycle time was 40~ms.


\subsection{Experimental Results}

With the setup proposed in Section~\ref{computer_setup} the learning capabilities within the framework were evaluated using D4PG to train two agents to solve the two problems. 

First, a different agent was trained on each of the environments using only experience gathered in the simulation.
After the training process was completed the resulting models were tested in simulation as well as in the real world environments. 
The trained agents were able to solve the real world environments with almost the same success rate achieved in simulation. 
As a consequence we show that the models trained in simulation can be deployed in the real world scenarios without any adaptation or further training needed. 
See the accompanying video
for an example of the performance of the two models.

\smallskip
\subsubsection{Results for Problem 1}
\begin{figure}[thpb]
    \centering
    \includegraphics[width=1\columnwidth]{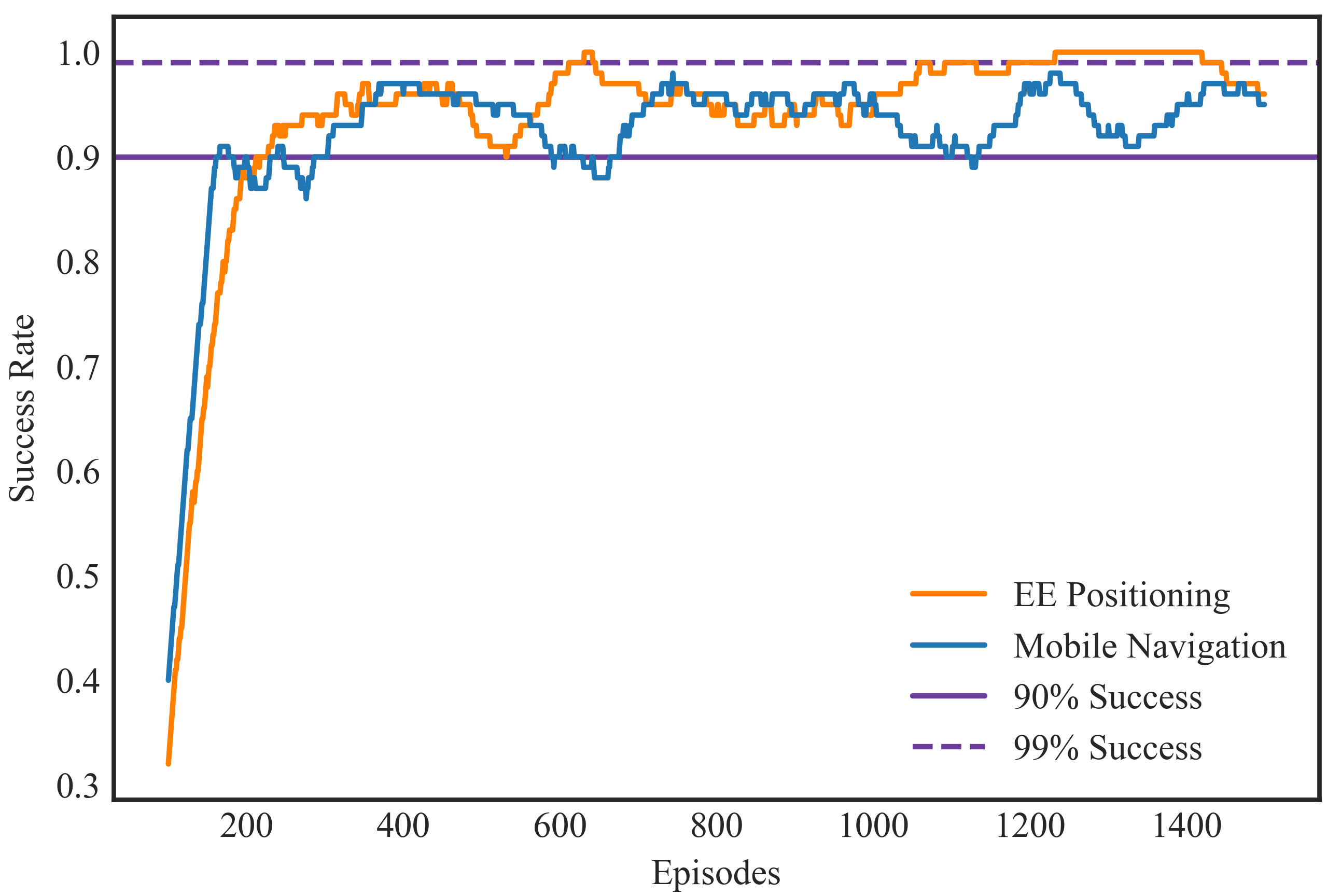}
    \caption{Development of the success rate while training for both problems. Success rate values are the ratio of successfully completed episodes over the last 100 consecutive episodes.}
    \label{fig:training_progress}
\end{figure}

The agent was trained for the task of mobile navigation with obstacle avoidance until the actors experienced about 4500 episodes each in the simulation environments. 
However, after the actors completed 400 iterations the success rate did not improve further and remained steady between 89 and 100 percent over 100 consecutive episodes (see Figure~\ref{fig:training_progress}).

For the final evaluation, the trained agent was tested for 100 episodes in the simulation environments as well as in the real world environments. 
In simulation the agent completed 93 episodes with success, 3 with collision with an obstacle and 4 times the agent could not reach the goal in time; resulting in a  93 \% success rate.
In the real world the agent completed 95 episodes with success and 5 with a collision with an obstacle; resulting in a 95 \% success rate.

\smallskip

\subsubsection{Results for Problem 2}
\begin{figure}[thpb]
    \centering
    \includegraphics[width=1\columnwidth]{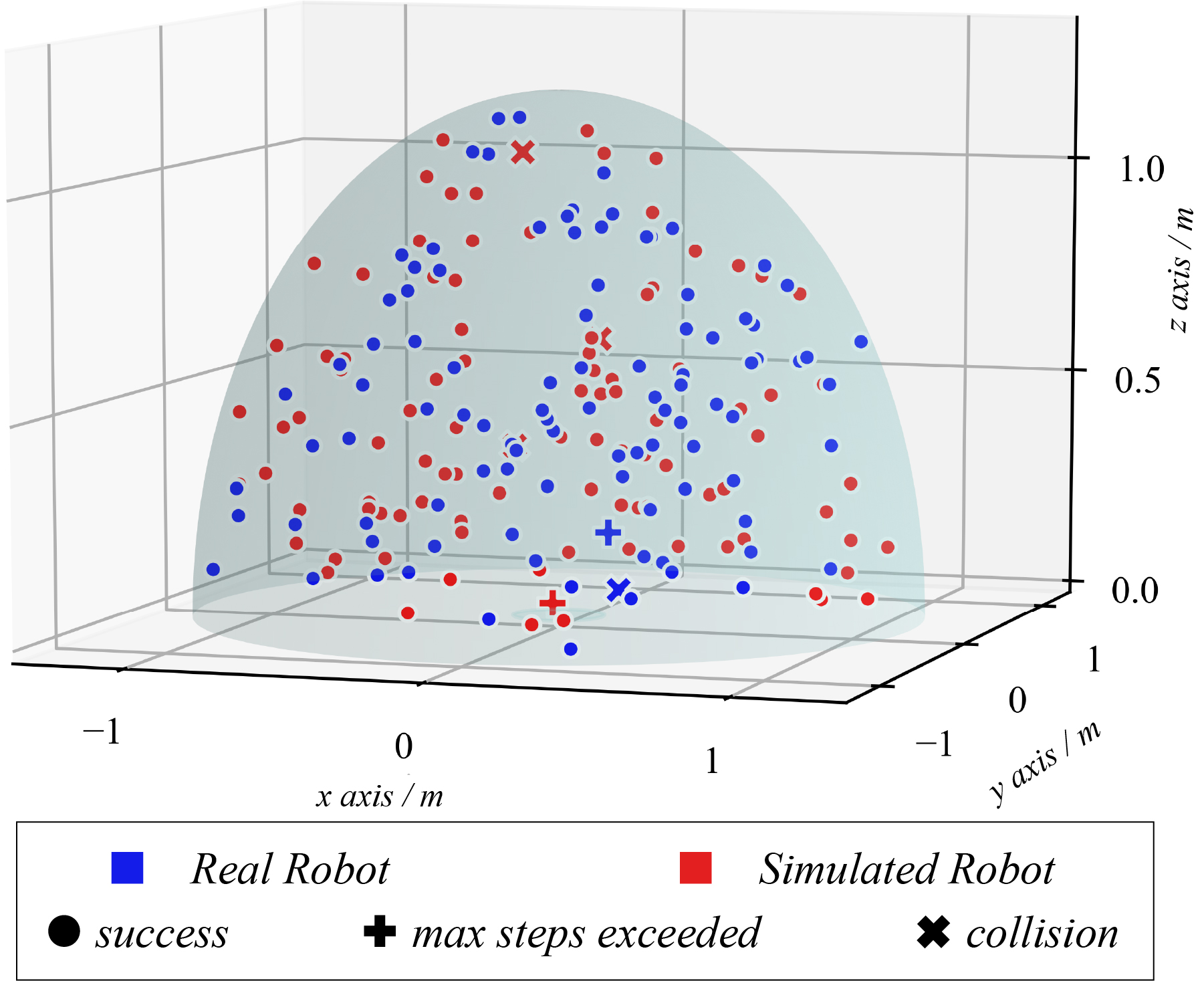}
    \caption{Shows all the generated goal positions and their corresponding terminal state for the end effector in the final evaluation of the positioning task. Goal positions are generated evenly across the working space of the UR10.}
    \label{fig:final_evaluation_ur10}
 \end{figure}

The results obtained for the end effector positioning task were very similar to \textit{Problem 1}.
The agent was trained for 5000 episodes for each actor in the simulated environment. 
Again, training converged earlier at around 600 episodes with steady success rates between 95 and 100 \% (see Figure~\ref{fig:training_progress}). 
The test was run for 100 episodes in both of the environments.
In simulation the agent completed 96 tasks with success, 3 with a collisions and once it could not reach the goal in time; resulting in a 96 \% success rate.
In the real world the agent completed 98 episodes with success, one with a self collision and once it exceeded the maximum number of steps; resulting in a 98 \% success rate. 
%
In Figure~\ref{fig:final_evaluation_ur10} we display the positions of the targets generated during the tests together with the results of the tasks.


\section{FRAMEWORK EVALUATION} \label{sec:framework_evaluation}

The selection of a framework or toolkit is not a trivial task; it is essential to understand the intended use and its limitations in advance.
To help the reader in the selection of a framework or toolkit, we describe and report in Table~\ref{tab:frameworks} a set of properties we found to be relevant for the use of DRL in robotics:

\begin{table*}[t]
\resizebox{\textwidth}{!}{
\begin{tabular}{lrcllcllclcl}
\toprule
\multicolumn{1}{c}{\multirow{2}{*}{\textbf{Framework}}} &
  \textbf{\begin{tabular}[c]{@{}l@{}}Community\\   Support\end{tabular}} &
  \phantom{a} &
  \multicolumn{2}{c}{\textbf{Diversity}} & 
  \phantom{a} &
  \multicolumn{2}{c}{\textbf{Heterogeneity}} &
  \phantom{a} &
  \textbf{Scalability} &
  \phantom{a} &
  \textbf{\begin{tabular}[c]{@{}l@{}}Software \\ Licensing\end{tabular}} \\ \cmidrule{4-5} \cmidrule{7-8}\addlinespace
\multicolumn{1}{c}{} &
  \multicolumn{1}{r}{\begin{tabular}[c]{@{}l@{}}\textit{Number} \\ \textit{of Forks}\end{tabular}} &&
  \multicolumn{1}{l}{\textit{Robots}} &
  \multicolumn{1}{r}{\begin{tabular}[c]{@{}r@{}}\textit{Number} \\ \textit{of Tasks}\end{tabular}} &&
  \multicolumn{1}{l}{\begin{tabular}[c]{@{}l@{}}\textit{Simulated}\\   \textit{Hardware} \\ \textit{Support}\end{tabular}} &
  \multicolumn{1}{l}{\begin{tabular}[c]{@{}l@{}}\textit{Real}\\   \textit{Hardware} \\ \textit{Support}\end{tabular}} &&
  \multicolumn{1}{l}{\begin{tabular}[c]{@{}l@{}}\textit{Distributed}\\   \textit{Hardware} \\ \textit{Support}\end{tabular}} &&
  \multicolumn{1}{l}{\begin{tabular}[c]{@{}l@{}}\textit{Simulation}\\   \textit{Platform}\end{tabular}} \\ \midrule

\multicolumn{1}{l}{\textit{OpenAI Gym - Robotics Suite}} &
  \multicolumn{1}{r}{5600} &&
  \multicolumn{1}{l}{\begin{tabular}[c]{@{}l@{}}Fetch Arm \\ Shadow Hand\end{tabular}} &
  \multicolumn{1}{r}{8} &&
  \multicolumn{1}{c}{Yes} &
  \multicolumn{1}{c}{No} &&
  \multicolumn{1}{c}{No} &&
  \multicolumn{1}{l}{MuJoCo} \\  \addlinespace
\multicolumn{1}{l}{\textit{DeepMind Control Suite}} &
  \multicolumn{1}{r}{291} &&
  \multicolumn{1}{l}{5  DOF Manipulator} &
  \multicolumn{1}{r}{5} &&
  \multicolumn{1}{c}{Yes} &
  \multicolumn{1}{c}{No} &&
  \multicolumn{1}{c}{No} &&
  \multicolumn{1}{l}{MuJoCo} \\  \addlinespace \addlinespace
\multicolumn{1}{l}{\textit{SURREAL Robotics Suite}} &
  \multicolumn{1}{r}{60} &&
  \multicolumn{1}{l}{Baxter} &
  \multicolumn{1}{r}{6} &&
  \multicolumn{1}{c}{Yes} &
  \multicolumn{1}{c}{No} &&
  \multicolumn{1}{c}{Yes} &&
  \multicolumn{1}{l}{MuJoCo} \\  \addlinespace \midrule
\multicolumn{1}{l}{\textit{RLBench}} &
  \multicolumn{1}{r}{30} &&
  \multicolumn{1}{l}{\begin{tabular}[c]{@{}l@{}}Franka Panda, Mico, \\ Jaco, Sawyer\end{tabular}} &
  \multicolumn{1}{r}{100} &&
  \multicolumn{1}{c}{Yes} &
  \multicolumn{1}{c}{No} &&
  \multicolumn{1}{c}{No} &&
  \multicolumn{1}{l}{\begin{tabular}[c]{@{}l@{}}Coppelia\\ Sim\end{tabular}} \\  \addlinespace
\multicolumn{1}{l}{\textit{SenseAct}} &
  \multicolumn{1}{r}{31} &&
  \multicolumn{1}{l}{\begin{tabular}[c]{@{}l@{}}UR5,  iRobot Create 2,\\ Dynamixel actuator\end{tabular}} &
  \multicolumn{1}{r}{5} &&
  \multicolumn{1}{c}{No} &
  \multicolumn{1}{c}{Yes} &&
  \multicolumn{1}{c}{No} &&
  \multicolumn{1}{l}{None} \\  \addlinespace
\multicolumn{1}{l}{\textit{gym-gazebo-2}$^4$} &
  \multicolumn{1}{r}{53} &&
  \multicolumn{1}{l}{MARA} &
  \multicolumn{1}{r}{6} &&
  \multicolumn{1}{c}{Yes} &
  \multicolumn{1}{c}{Yes} &&
  \multicolumn{1}{c}{No} &&
  \multicolumn{1}{l}{Gazebo} \\ \addlinespace \midrule \addlinespace
\textit{robo-gym} &
  \multicolumn{1}{r}{X} &&
  \multicolumn{1}{l}{MiR100,  UR10} &
  \multicolumn{1}{r}{2} &&
  \multicolumn{1}{c}{Yes} &
  \multicolumn{1}{c}{Yes} &&
  \multicolumn{1}{c}{Yes} &&
  \multicolumn{1}{l}{Gazebo} \\  \bottomrule
\end{tabular}
}
\caption{Table of comparison of DRL frameworks for robotic applications across the properties listed in Section~\ref{sec:framework_evaluation}.
}
\label{tab:frameworks}
\end{table*}

 \begin{enumerate}[label=\alph*)]
     \item \textbf{Community Support:} A good quality indicator for a framework is the level of adoption and support received from the community. As an indicator of that, we report the number of forks of the source code repositories. 
     \item \textbf{Diversity:} To help the development of more general AI it is crucial to test the algorithms on a diverse set of problems. This will be reported as the number of tasks and the robots included. 
     \item \textbf{Extensibility:}
     It is important to  facilitate the extension of a framework to different robots and sensors to allow research from other groups and companies.  
     It is challenging to objectively evaluate this property without first hand experience with the frameworks; as a consequence, we leave the assessment of this property to the reader. 
     \item \textbf{Heterogeneity:} Collecting experience from both real and simulated hardware and scenarios can be beneficial for the robustness of trained models. The support for real and simulated hardware is listed. 
     \item \textbf{Scalability:} Machine Learning algorithms require large quantities of data to train on. Being able to scale and parallelize data generation is fundamental to speed up the learning of new tasks. The capability of a framework to handle distributed hardware and software out of the box is outlined.   
     \item \textbf{Software Licensing:} Open source software has often accelerated research in multiple fields. It is important that not only the framework's code base but also the tools on which it relies are open source. Proprietary tools may have prohibitive costs that prevent researchers from engaging in the field. We report on which simulation platform each framework is based, while information on their licences is given in Section~\ref{physics_engines}.
     \item \textbf{Transferability:}  To accelerate the adoption of DRL techniques in real world scenarios, it is necessary to provide tools to simplify the transfer from simulated to real world. We specify whether a framework allows for this.
 \end{enumerate}


Table~\ref{tab:frameworks} shows that one of the current limitations of \textit{robo-gym} is the number of tasks implemented. 
On the other hand, it highlights that all the works aside from \textit{robo-gym} and gym-gazebo-2 are based on proprietary simulation software, and this poses obvious limitations.
In addition, it is shown that most of the existing frameworks only have support for simulated hardware, while SenseAct provides support for real hardware but not for simulated one, thus limiting the data collection capabilities. 
The only other framework that provides support for simulated and real hardware at the same time is gym-gazebo-2\footnote{The project is not active anymore}; however, this has been implemented only for the MARA robot, which is not produced anymore.  
Furthermore, \textit{robo-gym} provides multiple additional features:
\begin{itemize}
    \item integration of two commercially available industrial robots
    \item out of the box support for distributed hardware
    \item real and simulated robots interchangeability 
\end{itemize}

As a result \textit{robo-gym} is the most suitable option for developing DRL robotics applications that:
\begin{itemize}
    \item feature mobile robots and robot arms providing a ROS interface
    \item can be trained in simulation on distributed hardware
    \item can be transferred to real world use cases
    \item can be developed and trained without incurring in licensing costs
\end{itemize}

\smallskip
\section{CONCLUSION AND FUTURE WORK}
\label{sec:conclusions}

We introduced \textit{robo-gym}, the first open source and freely available framework that allows to train DRL control policies in distributed simulations and to apply them directly to the real world robots. 
The framework is built on open source software allowing the user to develop applications on own hardware and without incurring in cloud services fees or software licensing costs. 
The effectiveness of the framework has been proven with the development and evaluation of two industrial use cases featuring a mobile robot and a robot arm. 

This is the first necessary step towards the development of a tool chain that allows to develop new robot applications in simulation and to seamlessly transfer them to industrial scenarios.
Future efforts will go into extending the framework with new robot models and sensors and the integration of tools that simplify the implementation of increasingly complex problems.
The goal is to have a continuously growing toolkit that can serve as a solid base for developing research within the field of DRL in robotics.


\section*{APPENDIX}

The video attachment shows the experiments conducted in the lab demonstrating the control policies trained purely in simulation and directly deployed on the real robots. 

\section*{ACKNOWLEDGMENTS}
This research has received funding from the Austrian Ministry for Transport, Innovation and Technology (bmvit) within the project "Credible \& Safe Robot Systems (CredRoS)", from the "K\"arntner Wirtschaftsf\"orderung Fonds" (KWF) and the "European Regional Development Fund" (EFRE) within the CapSize project 26616/30969/44253, and the Austrian Forschungsförderungsfond (FFG) within the project “Flexible Intralogistics for Future Factories (FlexIFF)”.

We would like to thank Inka Brija\v{c}ak and Damir Mirkovi\v{c} for contributions in the early phase of the project. Furthermore, we would like to thank
Mathias Brandst\"{o}tter,
Bernhard Holzfeind,
Lukas Kaiser,
Barnaba Ubezio,
V\'{i}ctor M. Vilches, 
Matthias Weyrer and 
Lucas Wohlhart
for the useful discussions and the help in setting up the experiments.


\balance
\bibliographystyle{IEEEtran}
\bibliography{drl}


\end{document}